# Adaptive Parallel Iterative Deepening Search

**Diane J. Cook** COOK@CSE.UTA.EDU
*Department of Computer Science and Engineering*
*University of Texas at Arlington*
*Box 19015, Arlington, TX 76019 USA*

**R. Craig Varnell** CVARNELL@SFASU.EDU
*Department of Computer Science*
*Stephen F. Austin State University*
*Box 13063, Nacogdoches, TX 75962 USA*

## Abstract

Many of the artificial intelligence techniques developed to date rely on heuristic search through large spaces. Unfortunately, the size of these spaces and the corresponding computational effort reduce the applicability of otherwise novel and effective algorithms. A number of parallel and distributed approaches to search have considerably improved the performance of the search process.

Our goal is to develop an architecture that automatically selects parallel search strategies for optimal performance on a variety of search problems. In this paper we describe one such architecture realized in the EUREKA system, which combines the benefits of many different approaches to parallel heuristic search. Through empirical and theoretical analyses we observe that features of the problem space directly affect the choice of optimal parallel search strategy. We then employ machine learning techniques to select the optimal parallel search strategy for a given problem space. When a new search task is input to the system, EUREKA uses features describing the search space and the chosen architecture to automatically select the appropriate search strategy. EUREKA has been tested on a MIMD parallel processor, a distributed network of workstations, and a single workstation using multithreading. Results generated from fifteen puzzle problems, robot arm motion problems, artificial search spaces, and planning problems indicate that EUREKA outperforms any of the tested strategies used exclusively for all problem instances and is able to greatly reduce the search time for these applications.

## 1. Introduction

Because of the dependence AI techniques demonstrate upon heuristic search algorithms, researchers continually seek more efficient methods of searching through the large spaces created by these algorithms. Advances in parallel and distributed computing offer potentially large increases in performance to such compute-intensive tasks. In response, a number of parallel approaches have been developed to improve various search algorithms including depth-first search (Kumar & Rao, 1990), branch-and-bound search (Agrawal, Janakiram, & Mehrotra, 1988), A* (Evett, Hendler, Mahanti, & Nau, 1995; Mahapatra & Dutt, 1995), IDA* (Mahanti & Daniels, 1993; Powley, Ferguson, & Korf, 1993; Powley & Korf, 1991), and game tree search (Feldmann, Mysliwietz, & Monien, 1994), as well as to improve the run time of specific applications such as the fifteen puzzle problem (Kumar & Rao, 1990) and robot arm path planning (Challou, Gini, & Kumar, 1993). In addition to MIMD distributed-memory algorithms, parallel search algorithms have been developed for MIMD shared-memory systems (Kale & Saletore, 1990; Kumar & Rao, 1990) and SIMD architectures (Cook & Lyons, 1993; Evett et al., 1995; Karypis & Kumar, 1992; Mahanti & Daniels, 1993; Powley et al., 1993). While existing approaches to parallel search have many contributions to offer, comparing these approaches and determining the best use of each contribution is difficult because of the diverse search algorithms, implementation platforms, and applications reported in the literature.

In response to this problem, we have developed the Eureka parallel search engine that combines many of these approaches to parallel heuristic search. Eureka (Cook & Varnell, 1997) is a parallel IDA* search architecture that merges multiple approaches to task distribution, load balancing, and tree ordering, and can be run on a MIMD shared memory or distributed memory parallel processor, a distributed network of workstations, or a single machine with multithreading. Using our collection of search algorithms, we perform theoretical and empirical comparisons and observe that performance trends do exist as search space features are varied. To capitalize on these trends, Eureka uses a machine learning system to predict the optimal set of parallel search strategies for a given problem, which are then used to complete the search task.

## 2. Parallel Search Approaches

A number of researchers have explored methods for improving the efficiency of search using parallel hardware. We will focus in this paper on parallel search techniques that can be applied to IDA* search. IDA* performs a series of incrementally-deepening depth-first searches through the search space. In each iteration through the space, the depth of the search is controlled by an A* cost threshold. If a goal node is not found during a given iteration, search begins at the root node with a cost threshold set to the minimum $f(n)$ value in the search space that exceeded the previous threshold. IDA* is an admissible search algorithm which requires an amount of memory linear in the depth of the solution.

In this section we will review existing methods for parallelizing IDA* search. In particular, we will consider alternative techniques for task distribution, for dynamically balancing work between processors, and for changing the left-to-right order of the search tree.

### 2.1 Task Distribution

A search algorithm implemented on a parallel system requires a balanced division of work between contributing processors to reduce idle time and minimize redundant or wasted effort. One method of dividing up the work in IDA* search is with a parallel window search (PWS), introduced by Powley and Korf (1991). Using PWS, each processor is given a copy of the entire search tree and a unique cost threshold. The processors search the same tree to different thresholds simultaneously. If a processor completes an iteration without finding a solution, it is given a new unique threshold (deeper than any threshold yet searched) and begins a new search pass with the new threshold. When an optimal solution is desired, processors that find a goal node must remain idle until all processors with lower cost thresholds have completed their current iteration. A typical division of work using PWS is illustrated in Figure 1.

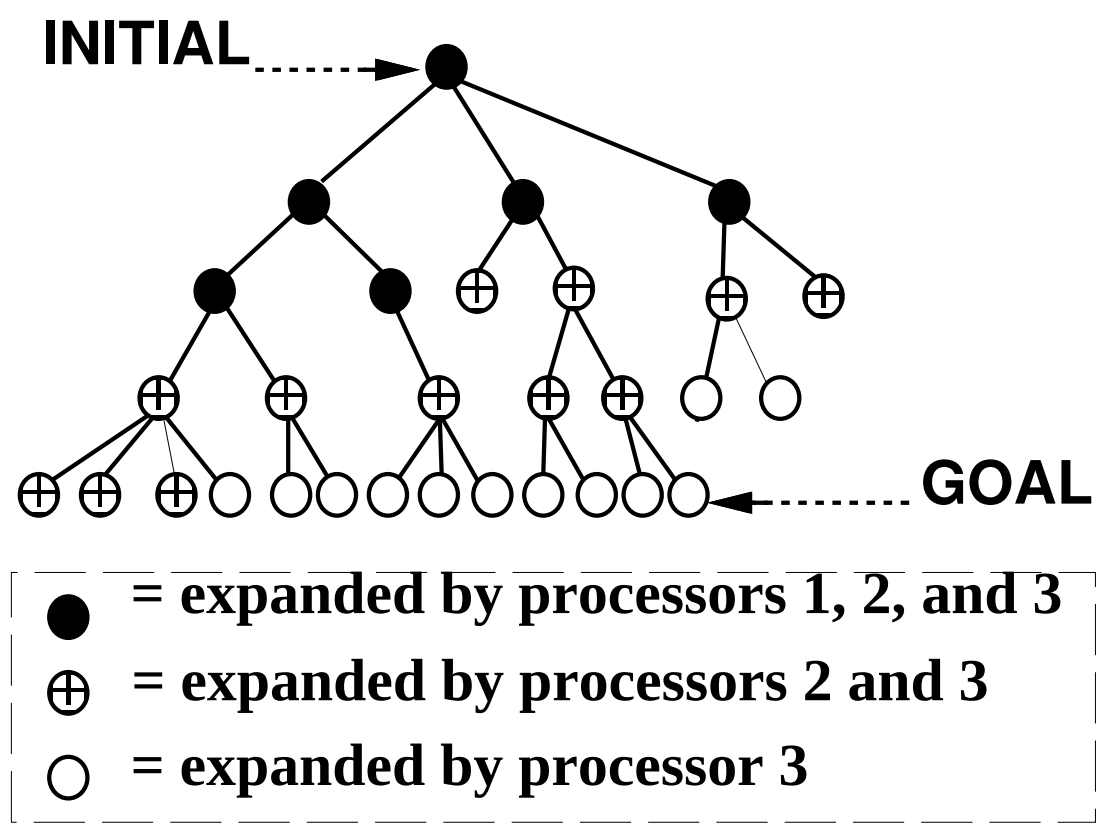

Figure 1: Division of work in parallel window search

One advantage of parallel window search is that the redundant search inherent in IDA* is not performed serially. During each non-initial iteration of IDA*, all of the nodes expanded in the previous iteration are expanded again. Using multiple processors, this redundant work is performed concurrently. A second advantage of parallel window search is the improved time in finding a first solution. If a search space holds many goal nodes, IDA* may find a deep solution much more quickly than an optimal solution. Parallel window search can take advantage of this type of search space. Processors that are searching beyond the optimal threshold may find a solution down the first branch they explore, and can return that solution long before other processors finish their iteration. This may result in superlinear speedup because the serial algorithm conservatively increments the cost threshold and does not look beyond the current threshold.

On the other hand, parallel window search can face a decline in efficiency when the number of processors is significantly greater than the number of iterations required to find an optimal (or a first) solution, causing all remaining processors to sit idle. This situation will occur when many processors are available, yet few iterations are required because the heuristic estimate is fairly accurate.

An alternative parallel search approach relies on distributing the tree among the processors (Kumar & Rao, 1990; Rao, Kumar, & Ramesh, 1987). With this approach, the root node of the search space is given to the first processor and other processors are assigned subtrees of that root node as they request work. As an alternative, the distributed tree search algorithm (DTS) employs breadth-first expansion until there are at least as many expanded leaf nodes as available processors. Processors receive unique nodes from the expanding process and are responsible for the entire subtree rooted at the received node. Communication-free versions of this distribution scheme have also been reported (Mahapatra & Dutt, 1995; Reinefeld & Schnecke, 1994). In all of these tree distribution approaches, the processors perform IDA* on their unique subtrees simultaneously. All processors search to the same threshold. After all processors have finished a single iteration, they begin a new search pass through the same set of subtrees using a larger threshold. A sample distribution of the search space is shown in Figure 2.

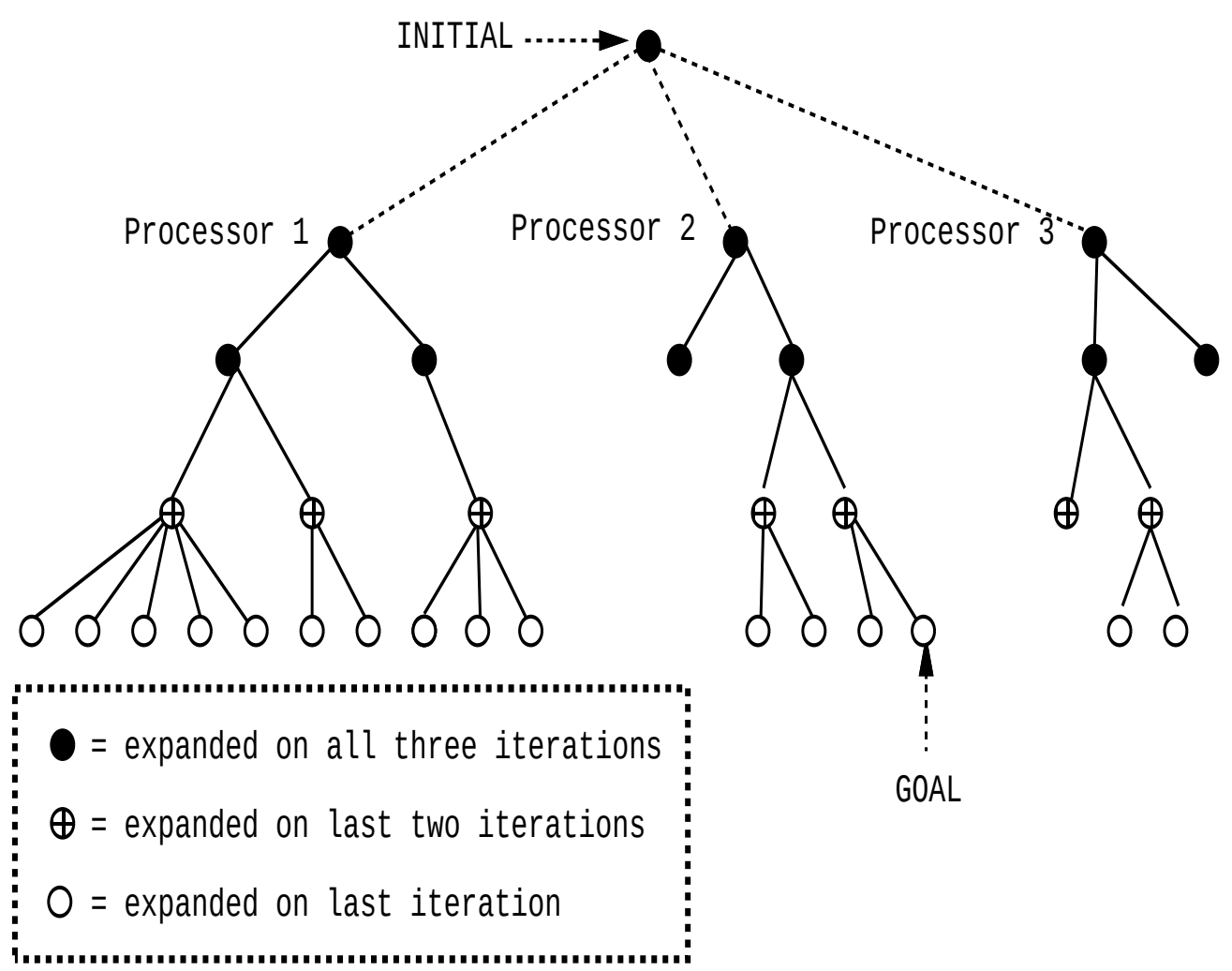

Figure 2: Division of work in distributed tree search

One advantage of this distribution scheme is that no processor is performing wasted work beyond the goal depth. Because the algorithm searches the space completely to one threshold before starting the search to a new threshold, none of the processors is ever searching at a level beyond the level of the optimal solution. It is possible, however, for DTS to perform wasted work at the goal depth. For example, in Figure 2 processor 3 searches nodes at the goal level that would not be searched in a serial search algorithm moving left-to-right through the tree.

A disadvantage of this approach is the fact that processors are often idle. To ensure optimality, a processor that quickly finishes one iteration must wait for all other processors to finish before starting the next iteration. This idle time can make the system very inefficient and reduce the performance of the search application. The efficiency of this approach can be improved by performing load balancing between neighboring processors working on the same iteration.

These described approaches offer unique benefits. Parallel window search is effective when many iterations of IDA* are required, when the tree is so imbalanced that DTS will require excessive load balancing, or when a deep, non-optimal solution is acceptable. On the other hand, dividing the search space among processors can be more effective when the branching factor is very large and the number of IDA* iterations is relatively small. A compromise between these approaches divides the set of processors into *clusters* (Cook, 1997). Each cluster is given a unique cost threshold, and the search space is divided between processors within each cluster, as shown in Figure 3. Setting the number of clusters to one simulates distributed tree search, and setting the number of clusters to the number of available processors simulates parallel window search.

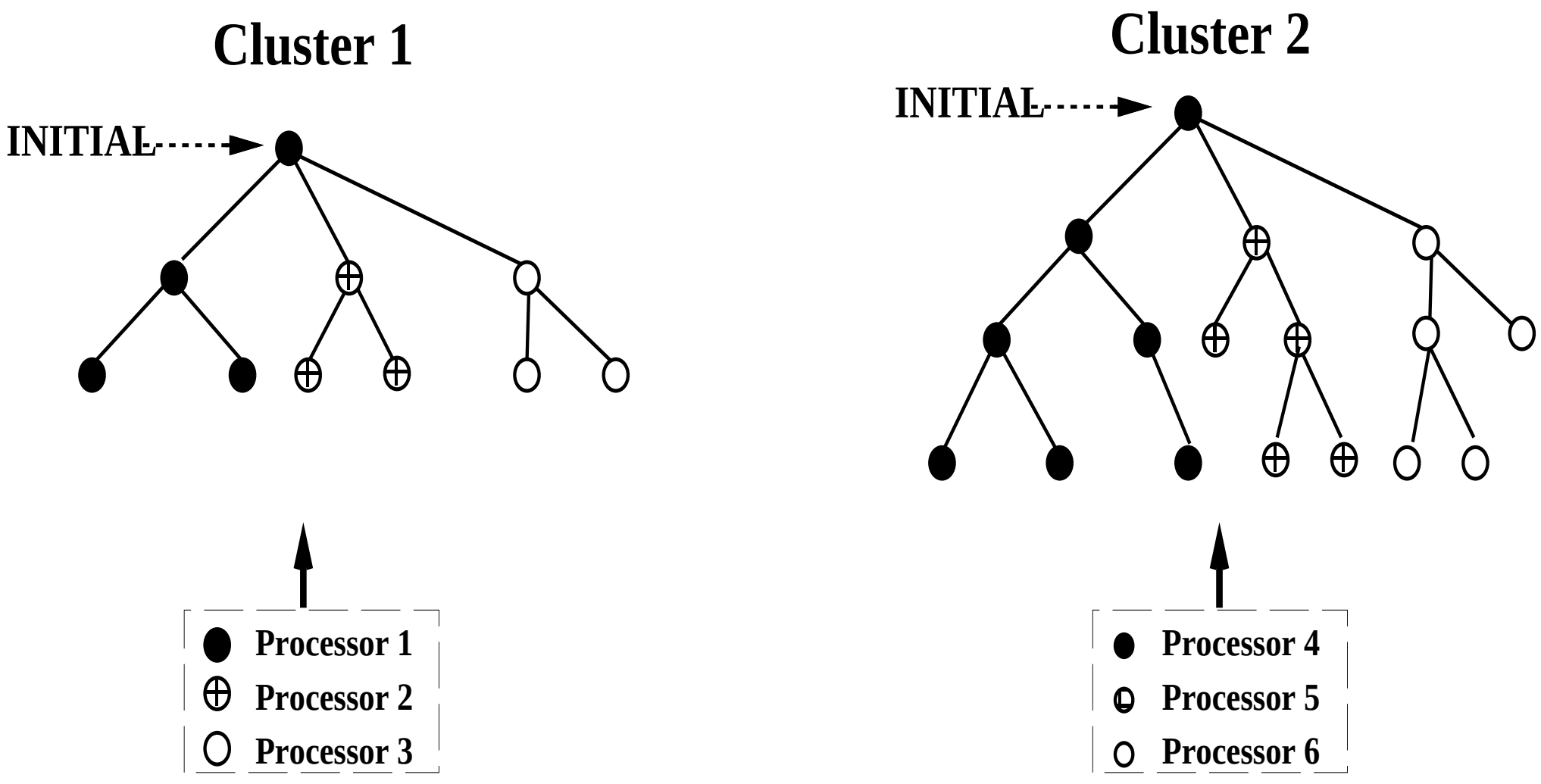

Figure 3: Space searched by two clusters, each with 3 processors

### 2.2 Load Balancing

When a problem is broken into disjoint subtasks the workload will likely vary among processors. Because one processor may run out of work before others, load balancing is used to activate the idle processor. The first phase of load balancing involves selecting a processor from which to request work. One example is the nearest neighbor approach (Mahapatra & Dutt, 1995); alternative approaches include selecting random processors or allowing a master processor to keep track of processor loads and send the ID of a heavily loaded processor to one that is idle. During the second phase of load balancing, the donating processor decides which work, if any, to give. A search algorithm typically stores nodes which have not been fully expanded on an Open list. When giving work to another processor, nodes can be given from the head of the list (deep in the tree), the tail (near the root), or from a sampling of all levels (Kumar & Rao, 1990).

A number of approaches have been introduced for reducing processor idle time using a load balancing operation. Using the quality equalizing strategy (Mahapatra & Dutt, 1995), processors anticipate idle time by sending out a work request when their load is almost empty, so that they can continue processing remaining nodes while waiting for a response. Alternative approaches are not receiver based, but allow an overly-loaded processor to initiate a load balance operation (Furuichi, Taki, & Ichyoshi, 1990; Rajpal & Kumar, 1993) or allow all processors to periodically shift work to keep the average load within acceptable bounds (Anderson & Chen, 1987; Saletore, 1990).

### 2.3 Tree Ordering

Problem solutions can exist anywhere in the search space. Using IDA* search, the children are expanded in a depth-first manner from left to right, bounded in depth by the cost threshold. If the solution lies on the right side of the tree, a far greater number of nodes must be expanded than if the solution lies on the left side of the tree. If information can be found to re-order the operators in the tree from one search iteration to the next, the performance of IDA* can be greatly improved.

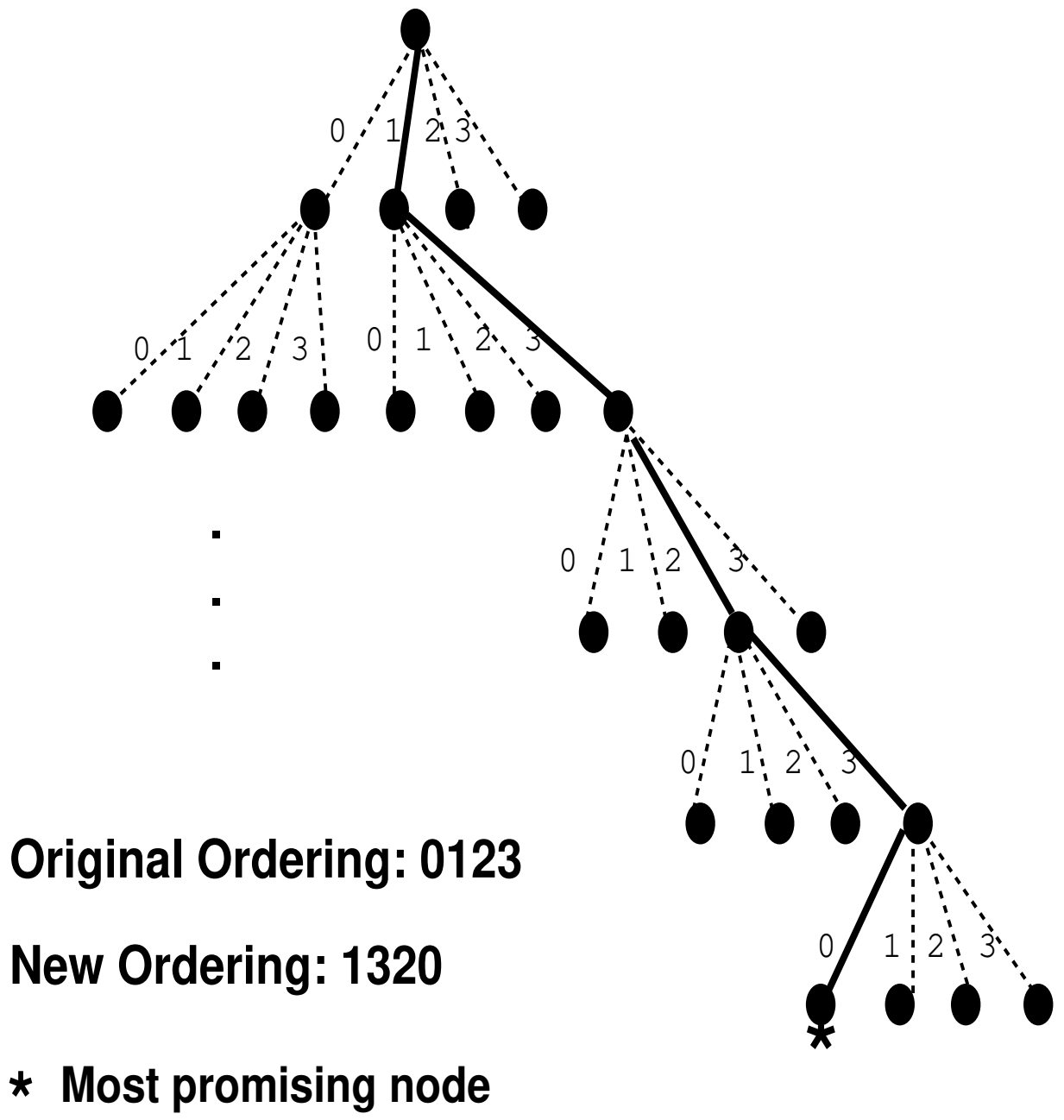

Figure 4: Operator ordering example

Powley and Korf suggest two methods of ordering the search space (1991). First, children of each node can be ordered and expanded by increasing heuristic distance to a goal node. Alternatively, the search algorithm can expand the tree a few levels and sort the *frontier set* (the set of nodes at that level in the tree) by increasing $h$ value. Search begins each iteration from the frontier set and this frontier set is updated each iteration. In both of these cases, although the nodes may have the same $f$ value, nodes with smaller $h$ values generally reflect a more accurate estimated distance and are preferred.

Instead of ordering individual nodes, Cook et al. (1993) order the set of operators to guide the next IDA* iteration to the *most promising node.* The most promising node is the node from the cut-off set (a child node not expanded in the previous iteration) with the smallest $h$ value. As an example, Figure 4 shows a search tree expanded using one iteration of IDA* with operator ordering 0, 1, 2, 3. The path to the most promising leaf node (indicated with a star) is 1 3 2 3 0. The new operator ordering is computed using the order of operators as they appear in this path after removing duplicates. Operators not appearing in the path are added to the end of the operator list, retaining their original relative ordering. Thus the ordering of operators for the example in Figure 4 changes from 0, 1, 2, 3 (try operator 0 first, operator 1 next, operator 2 next, and operator 3 last for every node in the tree) to 1, 3, 2, 0.

This section describes a number of alternative approaches to parallel search. Our theoretical empirical analyses in the following sections demonstrate that many of the reported approaches offer some benefit in certain conditions, but no single approach is always the most effective at scaling AI algorithms.

## 3. Comparison of Alternative Approaches to Parallel Search

One method of determining the comparative benefits of parallel search approaches is by determining the theoretical bounds on possible speedup obtained using each approach. A second method is to perform empirical comparisons between the approaches. In this section we will draw on theoretical analyses and empirical comparisons to determine where performance trends exist and to illustrate conditions under which alternative approaches can perform best.

### 3.1 Theoretical Analysis

In the literature we find theoretical analyses for the alternative approaches to one aspect of parallel search, namely task distribution. Kumar and Rao (1990) provide an analysis of the speedup of distributed tree search, and Powley and Korf (1991) provide an analysis of parallel window search. In this section we summarize these analyses with a unifying representation, and empirically compare the performance of the techniques using the derived equations.

These analyses assume that the average branching factor $b$ remains constant throughout the search space and that the least-cost goal is located at a depth $d$. We also let $b$ represent the heuristic branching factor, or the ratio of nodes generated during one iteration of IDA* to the number of nodes generated during the previous iteration of IDA*. Forcing the heuristic branching factor to be equal to the average branching allows the analysis to be the same as for incremental-deepening depth-first search.

For the distributed tree search analysis, we assume that a breadth-first expansion is used to generate enough nodes, $n$, to distribute one node to each of $P$ processors. Since $n = b^x$ and $n \geq P$, we can assume that the tree is expanded to depth $x$ where $x \geq log_b P$. We will assume a time of $2b^x \approx 2P$ to perform the node distribution and to collect the solution results from each processor. The speedup of distributed tree search, measured as the run time of the serial algorithm divided by the run time of the parallel algorithm, can be computed here as the number of serial nodes generated (assuming a constant node expansion time) divided by the number of serial node expansions performed by the parallel algorithm. This is derived in the literature (Kumar & Rao, 1990; Varnell, 1997) as

$$S = P\left(\frac{b^d + b^{d-1} + b^{d-2} + \cdots + b^2 + b}{b^d + b^{d-1} + b^{d-2} + \cdots + b^{x+1}}\right) + \frac{1}{2b^x}. \quad (1)$$

As $d$ increases, the leftmost fractional part of this equation approaches 1 and can be ignored. The $1/2b^x$ term contributes a minimal amount to the final value and can also be ignored. In this case, speedup approaches $P$, which represents linear speedup.

Figure 5 shows the performance of the distributed tree search algorithm based on these equations on a perfectly balanced tree and on a heavily imbalanced tree for $P = 10$, $b = 3$, and $d = 10$. In the imbalanced case, the size of the processors' search spaces varies as an exponential function where the first processor is assigned a majority of the work and the load decreases as the processor number increases. In this graph, the goal position ranges from the far left side of the tree (position = 0) to the far right side of the tree (position = 1). Performance of the search algorithm always peaks when the goal is on the far left side of a processor's portion of the search space. For the case of an imbalanced tree, much of the search space is assigned to a single processor, which increases the resulting amount of serial effort required.

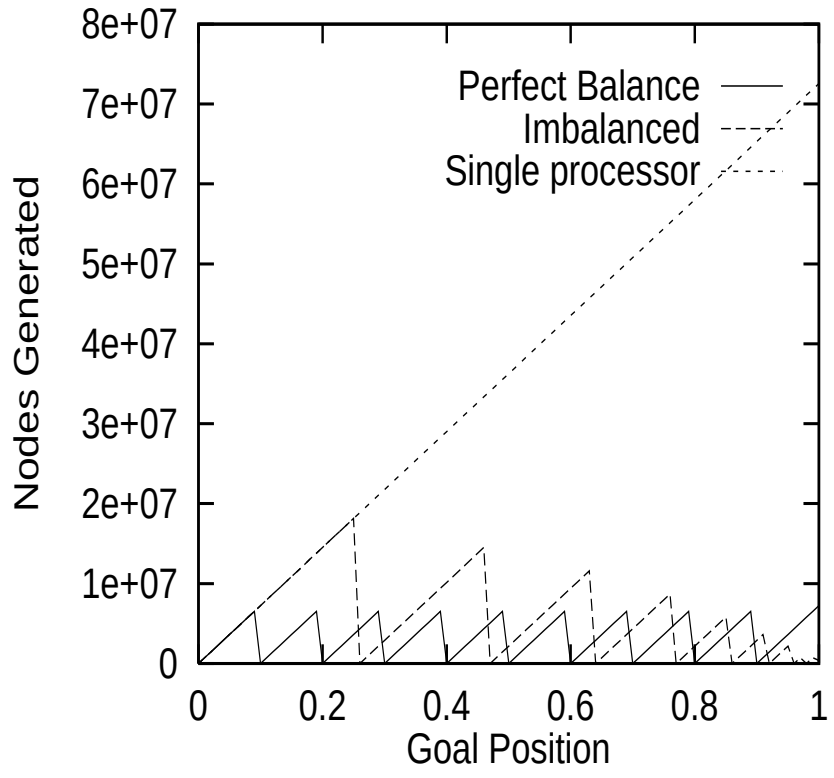

Figure 5: Distributed Tree Search Contrasting Tree Balance

We next consider the theoretical performance of the parallel window search algorithm. Recall that parallel window search operates by distributing the window sizes, or cost thresholds, to each available processor so each processor performs one iteration of IDA*. Since thresholds are not explored sequentially, the first solution found may not represent an optimal path. To ensure an optimal solution, all processors with a lower threshold must complete their current iteration of IDA*. In the worst case, this can make the performance of parallel window search equal to that of a serial version of IDA*.

In this analysis the assumption is made that a sufficient number of processors exists such that the goal iteration will start without delay. Speedup of parallel window search can be calculated as the ratio of the number of non-goal plus goal iteration nodes to the number of nodes generated by the processor performing the goal iteration. Powley and Korf generate this ratio using the notion of the left-to-right goal position $a$, defined as the fraction of the total number of nodes in the goal iteration that must be expanded before reaching the first goal node (1991). Speedup of parallel window search can thus be expressed as

$$S = \frac{b^{d-1}\left(\frac{b}{b-1}\right)^2 + ab^d\left(\frac{b}{b-1}\right)}{ab^d\left(\frac{b}{b-1}\right)} = 1 + \frac{1}{a(b-1)}. \quad (2)$$

Given this formula, we can empirically compare the performance of distributed tree search and parallel window search for $P = 10$, $b = 6$, and $d = 10$. From the graph in Figure 6 we can see that parallel window search will outperform distributed tree search only if the goal is located on the far left of the search space. We also observe that performance of distributed tree search peaks whenever the goal node is located on the far left of the subspace assigned to a particular processor.

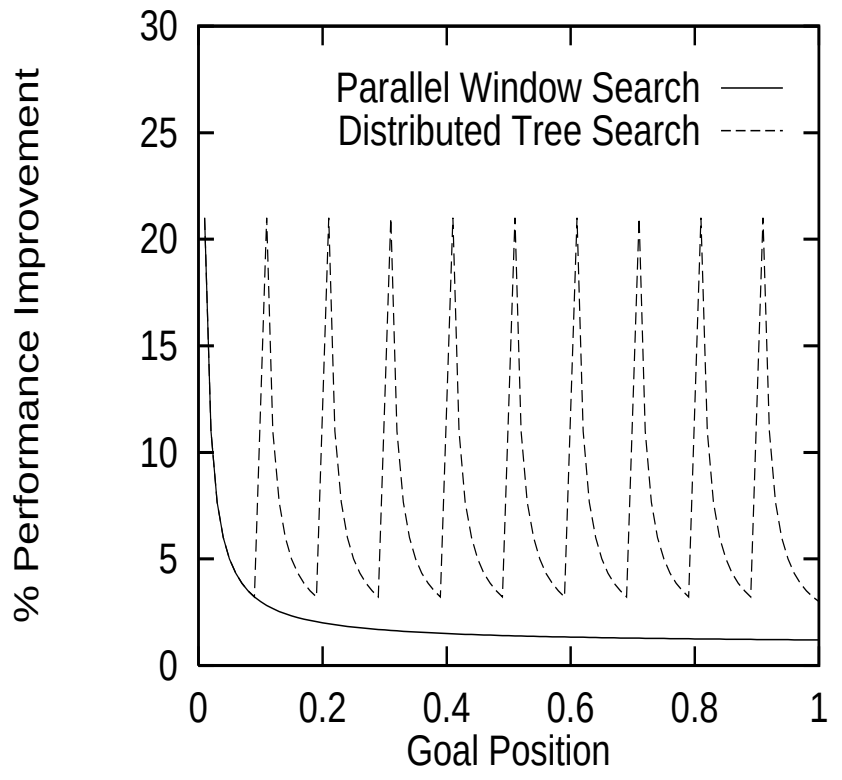

Figure 6: Distributed Tree Search vs. Parallel Window Search

Similar analyses have been provided to compare node ordering techniques and to determine the optimal number of clusters (Cook et al., 1993; Powley & Korf, 1991; Varnell, 1997). These analyses do indicate trends in the performance of alternative strategies based on features of the problem space, and can also be used to determine the theoretical performance of a particular technique for a given problem. However, the terms used to predict the performance in many of these analyses are not always measurable and many assumptions made are too constraining for real-world problems.

### 3.2 Empirical Analysis

A second method of determining the comparative benefits of parallel search approaches is via empirical analyses. We have implemented a number of the approaches to parallel search described earlier in this paper in the Eureka system. We have also constructed an artificial search space generator to provide a testbed for these experiments. Search space parameters can be established by the user, including:

- the cost of the optimal solution,
- the left-to-right position of the goal node in the space,
- the branching factor,
- the tree imbalance,
- the solution density (fraction of nodes at or beyond the optimal solution cost that represent goal nodes), and
- the heuristic error (the difference between the estimated and true distances to the nearest goal node).

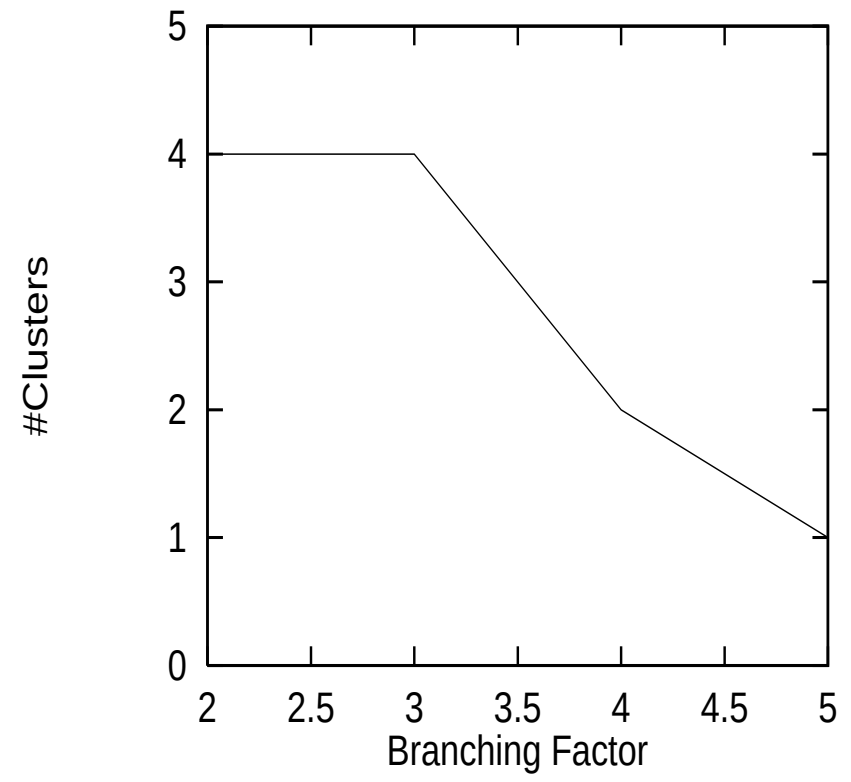

Figure 7: Branching Factor and Optimal Number of Clusters

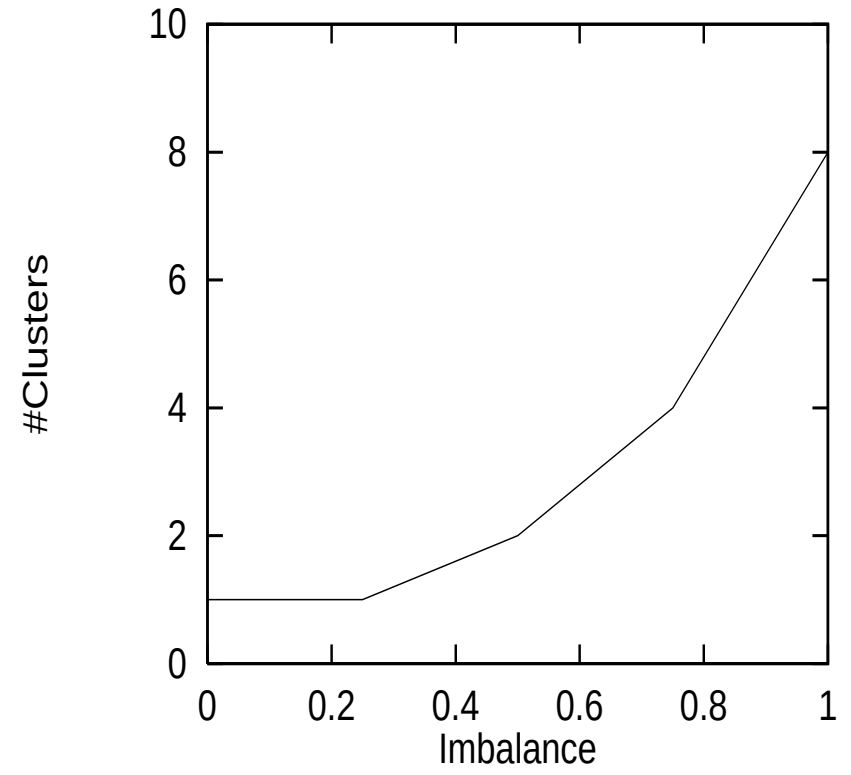

Figure 8: Tree Imbalance and Optimal Number of Clusters

All of the experiments described here were run on an nCUBE II message-passing multiprocessor using 32 processors.

In our first experiment we consider how the optimal number of clusters may be affected by features of the problem space including branching factor, tree imbalance, and solution position. Figures 7, 8, and 9 demonstrate that the optimal number of clusters increases as the branching factor decreases (with a balanced tree, an optimal cost of 16, and the goal on the far right side of the tree), as the imbalance increases (with a branching factor of 3, an optimal cost of 15, and the goal in the middle of the tree), or as the goal node moves to the right side of the tree (with a balanced tree, a branching factor of 3, and an optimal cost of 15). In no case did one single number of clusters always perform best.

In the next experiment we focus on the effects of operator ordering. Figure 10a demonstrates that employing operator ordering causes an increase in the optimal number of clusters, and Figure 10b shows that operator ordering (using perfect ordering information) results in a more significant improvement over no ordering as the solution node is positioned farther to the right in the tree. In this experiment the search trees are balanced with a branching factor of 4 and an optimal cost of 12. The dip in the plot occurs when the goal is positioned on the far left of a particular processor's subspace.

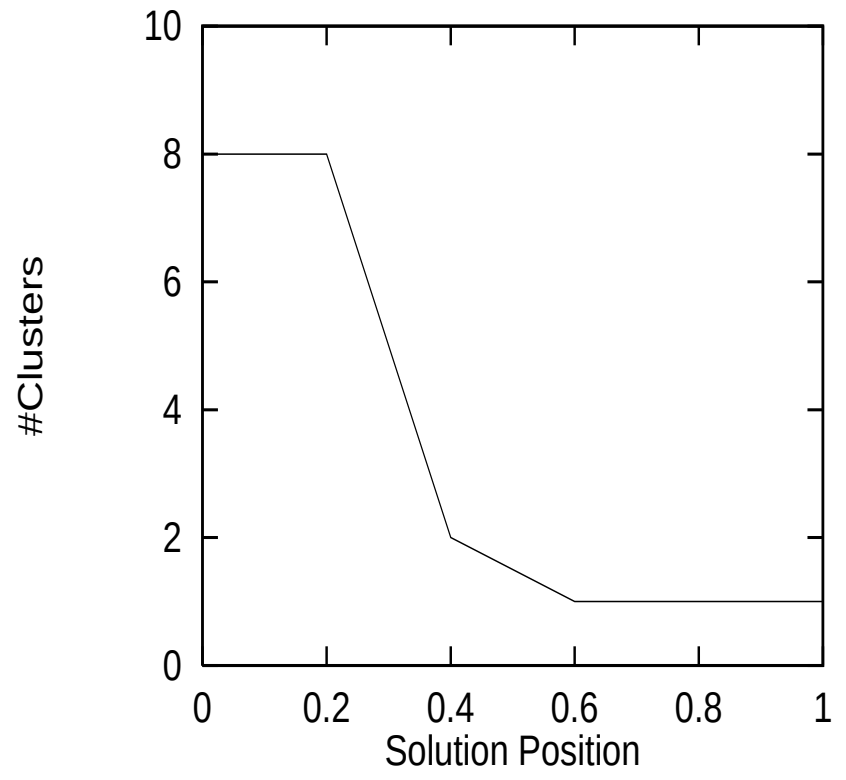

Figure 9: Goal Position and Optimal Number of Clusters

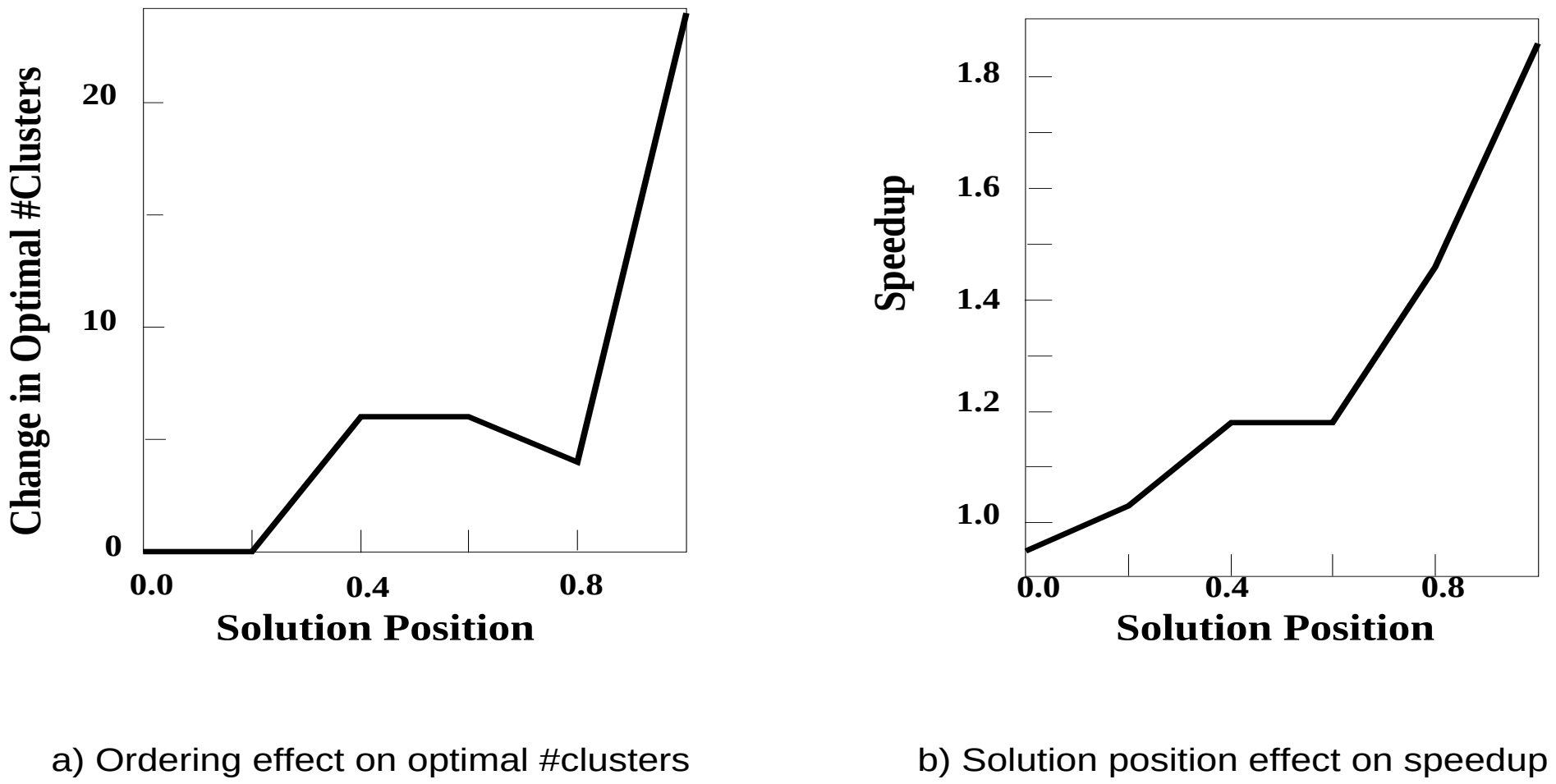

Figure 10: Ordering effects

## 4. The EUREKA System

The empirical and theoretical comparisons presented in the previous section indicate that alternative parallel search strategies perform well under different conditions, and that performance trends do exist which can be used to automatically select strategies and parameter settings for new problems. However, the results of these studies are not sufficient for automatically determining the appropriate set of strategies. One limitation is that information used to generate the formulas and to control the experiments, such as goal position, is not known in advance. Another limitation is that some of the assumptions, such as a constant branching factor, are not realistic for many applications. As a result, we need a method to automatically select optimal strategies for real-world problems given information that is available.

In response to this need, we add a machine learning component to the Eureka system. Eureka merges many of the approaches to parallel search discussed the previous section. Parameters can be set that control the task distribution strategy, the load balancing strategies, and the ordering techniques. In particular, the strategies that can be selected include:

- Distribution strategy [Kumar and Rao, Breadth-first]
- Number of clusters [1 ... #processors]
- Load balancing [On, Off]
- Processor selection [Neighbor, Random]
- Percentage of stack distributed upon request [0% ... 100%]
- Donating strategy [HeadOfList, TailOfList]
- Anticipatory load balancing trigger [0..stack_size]
- Ordering [Fixed, Local, TOIDA]

The user is allowed to determine the types of strategies to be used for a given problem. However, because such a decision is difficult to make without significant information about the search space, Eureka also has the capability of making all necessary strategy selections. Based on the characteristics of a search space, Eureka automatically configures itself to optimize performance on a particular application. Sample problems are fed as training examples to a machine learning system, which in turn learns the optimal strategies to apply to particular classes of search spaces. Applying machine learning to optimize software applications has been pursued in other areas of research. For example, Minton (1996) has applied a similar technique to automatically synthesize problem-specific versions of constraint-satisfaction algorithms. Research in other areas of computer science has yielded similar ideas of customizable environments applied to computer networks (Bhattacharjee, Calvert, & Zegura, 1997; Steenkiste, Fisher, & Zhang, 1997) and to interactive human-computer interfaces (Frank, Sukavirija, & Foley, 1995; Lieberman, 1998). This work is unique in allowing both problem-specific and architecture-specific features to influence the choice of strategies and in applying adaptive software techniques to parallel search. Eureka also offers a framework that can potentially automate both static and dynamic software customization.

To perform parallel search, Eureka executes the following steps:

1. Timings from sample problem instances are captured, varying each strategy parameter independently. In order to acquire an effective sample set, problems are selected from a variety of domains and parallel architectures.

2. For each problem instance, Eureka captures features of the problem space that are known to affect the optimal choice of strategy. These features include:

**Branching Factor ($b$):** The average branching factor of the search tree.

**Heuristic Error ($herror$):** The difference, on average, between the estimated distance to a goal node and the true distance to the closest goal node. This is estimated from the shallow search by computing the difference between the estimated distance to the goal at the beginning of the search and the smallest estimated distance to the goal for all leaf nodes of the shallow search, minus the actual distance from the root node to the leaf node.

**Imbalance ($imb$):** The degree to which nodes are unevenly distributed among subtrees in the search space.

**Goal Location ($loc$):** The left-to-right position of the first optimal solution node. This is estimated from the shallow search by determining which subtree contains nodes with the lowest estimated distances to the goal.

**Heuristic Branching Factor ($hbf$):** The ratio of nodes expanded between the current and previous IDA* iterations.

Features from the non-goal iterations represent attributes for the test case, and the strategy that results in the best performance (shortest run time) represents the correct classification of the problem instance for a given strategy choice.

3. Problem attributes are combined with the corresponding classes and are fed as training examples to a machine learning system. We use C4.5 (Quinlan, 1993) to induce a decision tree from the pre-classified cases. A rule base is generated for each concept to be learned, corresponding to each of the strategy decisions listed above that need to be made.

4. To solve a new problem, Eureka performs a shallow search through the space until roughly 200,000 nodes are expanded. If a goal is not found during the shallow search, the features of the tree are calculated at this point and used to index appropriate rules from the C4.5 database.

5. The learned rules recommend strategy choices given the features of the new problem space. Eureka then initiates a parallel search from the root of the space, employing the selected strategies. For many applications, the initial expansion takes only a few seconds and does not greatly affect the runtime of the search algorithm.

The described set of features and the amount of time to spend on the initial Eureka iteration are chosen based on our experimental data to yield the most helpful information in the shortest time. Searching enough iterations of the problem space until 200,000 nodes are generated takes less than 10 seconds on the problem domains we tested. Spending less time than this may yield erroneous information because features of the tree do not stabilize until several levels down in the tree. Searching additional iterations in general does not significantly improve the quality of information and the time requirements grow exponentially. Improved approaches may include performing the initial search until the stability of the space reaches a prescribed level, or periodically updating the C4.5 choices and adjusting the parallel search approaches dynamically during execution.

| | | Deviation Within Problems | | | | Deviation Between Problems | | | |
|---|---|---|---|---|---|---|---|---|---|
| Domain | Stat | Bf | Herror | Imb | Hbf | Bf | Herror | Imb | Hbf |
| 15puzzle | Std Dev | 0.000 | 1.748 | 0.001 | 2.839 | 0.002 | 2.577 | 0.008 | 4.889 |
| | Avg Value | 1.509 | 3.039 | 0.286 | 6.938 | 1.509 | 3.039 | 0.286 | 6.938 |
| RMP | Std Dev | 0.050 | 9.331 | 0.001 | 0.002 | 0.205 | 43.400 | 0.002 | 0.138 |
| | Avg Value | 2.643 | 11.265 | 0.148 | 1.105 | 2.643 | 11.265 | 0.148 | 1.105 |

Table 1: Problem Feature Value Deviations

Performing an initial shallow search has also been shown to be effective in other parallel search research. For example, Suttner (1997) performs parallel search of the first few IDA* iterations twice during parallel search in order to determine the number of individual tasks that should be distributed to each processor. Cook et al. (1993) also perform an initial shallow search in order to obtain more accurate operator ordering information to use in reordering the search space. In each case, the amount of time required to perform the extra initial search is minimal yet greatly improves the performance of the overall search task.

The selected features each demonstrate a significant influence on the optimal search strategy. Although feature values can change dramatically from one problem to the next, each feature remains fairly stable between levels of the same tree. As a result, computing the values of these features at a low level in the tree provides a good indication of the structure of the entire tree. The "Deviation Within Problems" entries in Table 1 show the standard deviation of each feature value over all search tree levels. The results are averaged over 98 problem instances in the fifteen puzzle domain[1] and 20 problem instances in the robot arm motion planning domain. Both of these test domains are described in the next section. The low values in these columns indicate that the features provide a good indicator of the structure of the problem space at all levels in the tree. In contrast, the "Deviation Between Problems" table entries show the standard deviation of the average feature value (averaged over all levels in each tree) over all problem spaces. The larger values in these columns indicate that the features can effectively distinguish between different problem spaces.

Eureka is written in C and is currently implemented on an nCUBE 2, on Linux workstations using the Parallel Virtual Machine (PVM) communication software, on a DEC Alpha using an implementation of Posix threads, on Windows NT running Java threads and Cilk threads, and as a distributed Java system using RMI. We are also currently investigating means of dynamically switching between strategies during execution as the problem structure or environment changes.

1. The two largest fifteen puzzle problem instances were not included due to the amount of time required to generate this data for these problems.

## 5. Experimental Results

In this section we will present experimental results that compare the performance of our adaptive parallel search strategy with each fixed strategy used exclusively for all problem instances. In particular, we will verify the following hypotheses in this section:

- Eureka's adaptive search techniques can be used to achieve speedup over a variety of applications, and can demonstrate improved results over using a fixed strategy for all problem instances.
- The adaptive search technique can employ training examples from multiple applications to improve overall performance. We will demonstrate this using testing and training examples combined from two application domains.
- The learning component of Eureka is able to significantly outperform any of the tested fixed strategies in terms of predicting the best strategy for a given problem instance.
- In addition to effectively making one strategy choice for a new problem, Eureka is most effective of all tested approaches at making all strategy decisions for a given problem instance.
- A variety of learning techniques can be used to assist in strategy selection and offer speedup over serial search, though performance will vary from one learning technique to another.

### 5.1 Test Domains

One of our test domains is the well-known fifteen puzzle problem. This problem consists of a $4 \times 4$ grid containing tiles numbered one to fifteen, and a single empty tile position called the *blank* tile. A tile can be moved into the blank position from an adjacent position, resulting in the four operators up, down, left, and right. Given the initial and goal configurations, the problem is to find a sequence of moves to reach the goal. A sample goal configuration is shown in Figure 11. The Manhattan Distance Function provides an admissible heuristic for this problem, and we use the 100 problem instances provided in Korf's test data (1991). [2]

Our second application domain is the robot arm motion planning problem. Traditional motion planning methods are very costly when applied to a robot arm, because each joint has an infinite number of angles to which it can move from a given configuration, and because collision checking must be performed for each arm segment. Craig provides a detailed description of the calculations necessary to determine the position of the end effector in the 3D workspace given the current joint angles (1989). For our experiments, we use the parameters defined for the Puma 560 robot arm with six degrees of freedom shown in Figure 12. The size and layout of the room is the same for each of our test problems, but we vary the initial and goal arm configurations to generate 20 problem instances. The

2. Because of time constraints, we estimated the serial run time for the two largest fifteen puzzle problem instances using the number of required node expansions for these problems published in Korf's paper and the mean node expansion time averaged over the next five largest problem instances.

| | | | |
|---|---|---|---|
| | 1 | 2 | 3 |
| 4 | 5 | 6 | 7 |
| 8 | 9 | 10 | 11 |
| 12 | 13 | 14 | 15 |

Figure 11: Fifteen puzzle problem instance

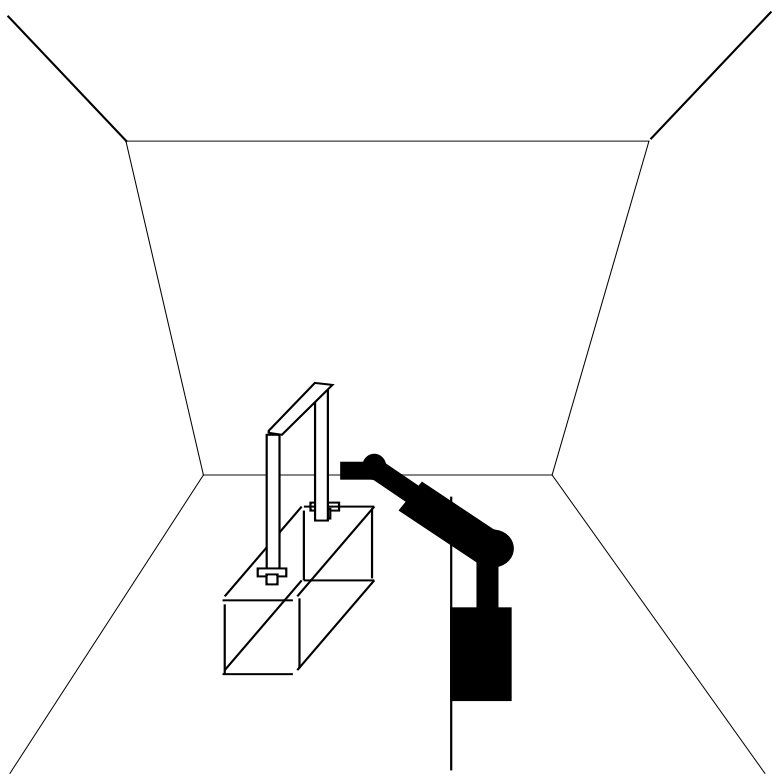

Figure 12: Robot arm motion planning problem instance

robot arm path planning problem is particularly difficult because considering every possible arm movement results in a search space with an infinite branching factor. We encode one possible move size for each joint, resulting in a branching factor of 6. The resolution of the moves can be determined by the user, and for these experiments we choose a resolution of 1 degree.

Our third test domain uses the artificial search space in which parameters including branching factor, tree imbalance, solution cost, heuristic error, and left-to-right goal position can be specified by the user. We generate 20 problem instances for use in the experiments.

For our fourth test domain, we integrate our own C-based version of the SNLP nonlinear planner (Barrett & Weld, 1992) into Eureka. To conform with the Eureka architecture, the integrated planner utilizes IDA* search instead of the Best-First search method employed by SNLP. Each plan repair step (filling an open condition or handling a threat) is treated as a separate node in the search space to be explored. We compute the cost of a plan solution as the number of operations and constraints in the plan, and the distance to the goal is estimated using the number of remaining flaws. We select 20 problem instances for our experiments from the blocks-world, Towers of Hanoi, and monkey-and-bananas planning domains.

To create test cases, we run each problem instance multiple times, once for each parallel search strategy in isolation. The search strategy that produces the best speedup is considered to be the "correct" classification of the corresponding search tree for C4.5 to learn. Test cases are run on 64 processors of an nCUBE 2 and on 8 distributed workstations using

| Approach | 15Puzzle | Fil-15P | RMP | Fil-RMP |
|---|---|---|---|---|
| Kumar and Rao | 52.02 | 65.89 | 63.10 | 57.46 |
| Breath-first | **53.03** | 65.82 | 59.77 | 55.90 |
| C4.5 | 52.31 | **79.17** | **64.15** | **61.08** |
| Combined-C4.5 | 61.30 | | | |

Table 2: Distributed Tree Search Speedup Results

PVM. In the first set of experiments we fix all strategy choices but one and vary the strategy choice under consideration. The default parameter choices are:

- Distribution strategy = Distributed Tree Search
- Number of clusters = 1
- Load balancing = On
- Processor selection = Neighbor
- Percentage of stack distributed upon request = 30%
- Donating strategy = TailOfList
- Anticipatory load balancing trigger = 0
- Ordering = Fixed

We compare the results of C4.5-selected strategies to each strategy used exclusively for all problem instances. Speedup results for various strategy decisions averaged over all problem instances are shown in the following sections. The best average speedup is highlighted in each case. The C4.5 results are captured by performing a ten-fold cross validation on the fifteen puzzle data and a three-fold cross validation on the robot arm motion planning, artificial, and nonlinear planning data sets. Specifically, a decision tree is created from the training cases and used to select the strategy for the test cases. C4.5 speedup is averaged over all test cases for one iteration of this process, and the final values are averaged over cross-validation iterations.

### 5.2 Distribution Results

Test results are obtained for the two alternative distribution approaches. The fifteen puzzle, the robot arm motion planning problem, and the artificial search space serve as problem domains. Experimental results are shown in Table 2. For each domain, the control variable is indicated under the “Approach” column.

The breadth-first distribution performs slightly better than C4.5 for the fifteen puzzle data set; however, the C4.5 recommendations outperform the breadth-first approach for the robot motion problem domain. The row labeled *Combined-C4.5* is the result of merging the fifteen puzzle results with the robot motion planning results and running the combined data set through C4.5. The performance is better than for the fifteen puzzle but slightly worse than for the robot motion planning domain.

| Stat | Small | Medium | Large |
|---|---|---|---|
| Avg Coef Var | 1.32 | 8.82 | 9.95 |
| Avg Speedup | 7.23 | 52.40 | 54.09 |

Table 3: Average Speedup Standard Deviation

Note that using the filtered 15 puzzle data, Eureka achieves a speedup of 79.17 although the number of processors used is only 64. These parallel search algorithms can produce superlinear speedup (speedup greater than the number of processors) because the parallel algorithms do not completely imitate the serial algorithm. For example, using distributed tree search individual subtrees are assigned to separate processors. If a goal node is located on the far left side of the rightmost subtree in the search space, the processor searching this subtree will quickly find the goal node, thus terminating search after only a few node expansions. In contrast, the serial algorithm in a left-to-right search will completely search all other subtrees to the cost threshold before searching and finding the goal node in the rightmost subtree. Thus the serial algorithm will perform disproportionately more search than all processors combined using the parallel algorithm. Each type of parallel search approach described in this paper can yield superlinear speedup under certain conditions. Some algorithms more closely imitate serial search, but at a potential loss of overall performance (Kale & Saletore, 1990).

Eureka's selection of strategies in the fifteen puzzle domain does not perform consistently better than using some of the strategies in isolation. One reason for this disappointing performance is the nature of the training data. Although we use the strategy that achieves the best run time as the correct "classification" for a given problem instance, there does not always exist a clear winner for each problem instance. On some problem instances one strategy dramatically outperforms the others. On other problem instances two or more strategy selections perform almost equally well.

This problem is exacerbated by the fact that there is some noise inherent in the collected run times. To demonstrate the amount of error that can be present in the timings we select twelve instances of the fifteen puzzle problem (four small, four medium, and four large instances), and time five runs of each instance with identical strategy parameters on an nCUBE 2. We compute the standard deviation of the speedups for five runs of the same problem instance, and then divide the result by the sample mean to ensure the result is not affected by the magnitude of the speedup values. This coefficient of variation averaged over all problem instances in the category is listed in Table 3 along with the average speedup for the instances in the problem category. As shown, the amount of error present in the timings can be quite large, and when two strategies perform almost equally well, the winner for any given run can be almost arbitrary.

To account for such misleading data, we sort all problem instances in the fifteen puzzle domain by the amount of variance of the strategy timing results. Those problem instances that yield a clear strategy winner are placed at the top of the list. We then filter the data set to keep only the top third of the sorted problem instances. The instances in the top third of this filtered data set are duplicated in the training set. The results of Eureka's performance on this filtered training set is shown for each experiment in the Fil-15P column. We perform a similar filtering step to the robot arm motion planning domain data.

| Approach | 15Puzzle | Fil-15P | RMP | Fil-RMP |
|---|---|---|---|---|
| Kumar & Rao | .6033 (.07) | .5809 (.00) | .5079 (.34) | .2000 (.03) |
| Breadth-first | .3967 (.14) | .4190 (.01) | .4921 (.11) | .8000 (.01) |
| C4.5 | .4533 | .1619 | .6429 | .0667 |

Table 4: Distributed Tree Search Classification Results

This approach can be used in each domain in which problem instances do not always yield a clear strategy winner. For example, problem instances drawn from the planning domain and artificial domain all demonstrate high variance of strategy timings, thus all problem instances are utilized. For any given domain, the number and type of test cases to use as training data can be selected based on the amount of variance of strategy results. The disadvantage of the filtered-data method is that cases in which two strategies yield similar timings may be discarded, even when the two strategies perform much better than other possible strategies.

The results in Table 3 verify that Eureka can automatically select parallel search strategies that yield greater speedup than using any single strategy for all problem instances pulled from the filtered data sets. However, this table does not indicate how well the machine learning component is performing at the classification task. One danger in listing only speedup results is that the numbers may be biased by the magnitude of several large problem instances in which Eureka correctly picks the best strategy.

In Table 4 we measure how well the system classifies each new test problem. Once again we perform ten-fold cross validation and show mean classification error for each approach. The fixed strategies (Kumar and Rao, Breadth-first) always pick the correct classification of a problem instance as their own, and C4.5 uses its decision tree to pick the classification of the problem instance. Significance values are gathered using a paired student t-test and are shown in parentheses following the mean error. In each of the filtered data sets, C4.5 significantly outperforms either fixed approach ($p \leq 0.03$) when predicting the correct classification of unseen problem instances.

### 5.3 Clustering Results

In this experiment Eureka selects the optimal number of clusters to use for each problem instance. By combining the features of distributed tree search and parallel window search, it is possible to achieve better performance than when each approach is used in isolation.

The clustering algorithm is tested using 1, 2, and 4 clusters on 64 processors of an nCUBE 2 for the fifteen puzzle, robot motion planning, and SNLP domains, and using 1 and 2 clusters for the fifteen puzzle domain on a distributed network of 8 PCs. Test results for the clustering algorithm are presented in Table 5.

| Approach | 15Puzzle | Fil-15P | RMP | Fil-RMP | Planning | PVM-15P |
|---|---|---|---|---|---|---|
| 1 Cluster | 52.02 | 65.21 | 60.01 | 62.75 | 108.46 | 7.69 |
| 2 Clusters | 57.04 | 64.97 | 80.09 | 79.98 | 145.0 | 7.03 |
| 4 Clusters | 56.83 | 49.57 | **162.86** | 153.17 | 129.15 | — |
| C4.5-nCUBE | **58.90** | **86.76** | 126.32 | **164.96** | **195.19** | **7.72** |
| Combined-C4.5 | 73.90 | | | | — | |

Table 5: Clustering Speedup Results

| Approach | 15Puzzle | Fil-15P | RMP | Fil-RMP |
|---|---|---|---|---|
| 1 Cluster | .5556 (.04) | .4595 (.00) | .7540 (.11) | .7778 (.01) |
| 2 Clusters | .6956 (.32) | .6738 (.00) | .9048 (.04) | .7778 (.02) |
| 4 Clusters | .7489 (.18) | .8667 (.00) | .3413 (.11) | .4444 (.01) |
| C4.5 | .6756 | .2333 | .4921 | .1481 |

Table 6: Clustering Classification Results

Table 5 demonstrates that Eureka's automatic strategy selection using C4.5 outperforms any fixed strategy in almost all domains, and always performs best when the filtered data sets are used. The table also indicates that the optimal number of clusters on average varies from one domain to another, thus reinforcing the need for automatic selection of this parameter. In the PVM experiments, because only eight processors are available we experimented with 1 or 2 clusters for each problem instance. The combined results are again collected from the test cases for the fifteen puzzle and robot arm motion planning domains.

The classification results for choice of number of clusters are shown in Table 6. On the filtered data set, C4.5 outperforms all fixed strategies at a significance level of $p \leq 0.02$.

### 5.4 Ordering Results

In this experiment we demonstrate Eureka's ability to pick a method of ordering the tree for expansion. Table 7 shows the results of this experiment. For the fifteen puzzle, the two tested fixed orderings are (Up, Left, Right, Down) and (Down, Left, Right, Up). For the robot arm motion planning domain, two fixed orderings are tested corresponding to ordering joint moves from the base of the arm to the end effector, and ordering joint moves from the end effector first down to the base of the arm last. Only one ordering is used for the artificial domain.

In this experiment, C4.5 yields the best speedup results for all databases, filtered or unfiltered. In the artificial domain, because perfect ordering information is available the TOIDA strategy also yields the best possible speedup results. The combined results are generated using the fifteen puzzle and robot arm motion planning problem instances.

Table 8 shows the results of classifying ordering problems on the filtered and unfiltered data sets. While C4.5 always yields the best average speedup, the learning system does not yield the best classification accuracy on unfiltered data, though it does achieve the best results on the filtered data sets. On the filtered data sets, C4.5 outperforms fixed strategies at a significance value of $p \leq 0.02$ or better.

| Approach | 15Puzzle | Fil-15P | RMP | Fil-15P | Artificial |
|---|---|---|---|---|---|
| Ordering 1 | 49.58 | 49.58 | 64.79 | 58.42 | 59.22 |
| Ordering 2 | 52.02 | 65.41 | 65.41 | 80.05 | — |
| TOIDA | 50.77 | 66.13 | 73.30 | 79.30 | **61.03** |
| Local | 50.75 | 68.92 | 75.37 | 82.25 | 60.62 |
| C4.5 | **50.97** | **123.79** | **78.52** | **87.28** | **61.03** |
| Combined-C4.5 | 55.28 | | | | — |

Table 7: Ordering Speedup Results

| Approach | 15Puzzle | Fil-15P | RMP | Fil-RMP |
|---|---|---|---|---|
| Fixed | .6972 (.44) | .7667 (.00) | 1.000 (.01) | 1.000 (.00) |
| TOIDA | .6194 (.07) | .7167 (.00) | .3016 (.21) | .3000 (.02) |
| Local | .6833 (.34) | .5167 (.00) | .6984 (.02) | .7000 (.00) |
| C4.5 | .7069 | .1429 | .4444 | .0000 |

Table 8: Ordering Classification Results

### 5.5 Load Balancing Results

Load balancing significantly affects the performance of a majority of parallel algorithms. When work is divided evenly among processors, no load balancing is necessary. Heuristic search frequently creates highly irregular search spaces, which results in load imbalance between processors. Eureka permits load balancing operations during iterations of IDA*. A processor with nodes available on its open list may donate some or all of the nodes to a requesting processor. Decisions that affect system performance include deciding when to balance the load, identifying a processor from which to request work, and deciding how much work to donate.

In the first load balancing experiment we test Eureka's ability to select the appropriate processor polling strategy. We have implemented the asynchronous round robin and the random polling approaches. On the nCUBE 2, a processor's $D$ neighbors are polled for work (using the nCUBE's hypercube topology, $D$ corresponds to $log_2 P$) whereas in the PVM environment, a processor's right and left neighbors are polled ($D = 2$ because the workstations are connected with a ring topology). The results of this experiment are listed in Table 9.

Table 9 shows that once again C4.5 yields the best speedup in most cases and always yields the best speedup on filtered data sets. Among the fixed results, no single approach outperforms the others on all data sets.

| Approach | 15Puzzle | Fil-15P | RMP | Fil-RMP |
|---|---|---|---|---|
| Neighbor | 52.02 | 65.21 | 58.81 | 57.67 |
| Random | **55.35** | 70.75 | 58.17 | 56.03 |
| C4.5 | 50.55 | **75.01** | **61.31** | **60.84** |
| Combined-C4.5 | 56.71 | | | |

Table 9: Load Balancing Speedup Results

| Approach | 15Puzzle | Fil-15P | RMP | Fil-RMP |
|---|---|---|---|---|
| Neighbor | .5306 (.08) | .5762 (.00) | .2937 (.21) | .2000 (.04) |
| Random | .4694 (.03) | .4238 (.00) | .7063 (.03) | .8000 (.00) |
| C4.5 | .3806 | .1429 | .4048 | .0000 |

Table 10: Load Balancing Classification Results

Table 10 summarizes the classification results of the fixed strategies in comparison to the C4.5 classifications. For each of the filtered data sets, C4.5 outperforms any fixed strategy with a significance of p$\leq$0.04 or better.

The second load balancing experiment demonstrates Eureka's ability to determine the optimal amount of work to donate upon request. If too little work is donated, the requesting processor will soon return for more work. If too much work is donated, the granting processor will soon be in danger of becoming idle. Table 11 lists the results of this experiment, demonstrating once again that the learning system is capable of effectively selecting load balancing strategies, except when the unfiltered test cases from the fifteen puzzle are used (on the nCUBE and on the distributed network of workstations). The combined results are generated using training cases from the fifteen puzzle and robot arm motion planning nCUBE examples.

Table 12 lists the classification accuracy results. C4.5 does not perform significantly better than the fixed strategies for the unfiltered data, but does perform significantly better (p$\leq$0.04) than the fixed strategies for the filtered data.

| Approach | 15Puzzle | Fil-15P | RMP | Fil-RMP | PVM-15P |
|---|---|---|---|---|---|
| 30%-nCUBE | 52.02 | 65.21 | 63.10 | 55.49 | **7.69** |
| 50%-nCUBE | **53.68** | 61.95 | 61.26 | 60.13 | 7.49 |
| C4.5-nCUBE | 51.28 | **76.35** | **63.67** | **62.13** | 7.50 |
| Combined-C4.5 | 55.44 | | | | — |

Table 11: Distribution Amount Speedup Results

| Approach | 15Puzzle | Fil-15P | RMP | Fil-RMP |
|---|---|---|---|---|
| 30% | .5639 (.26) | .5333 (.00) | .1984 (.00) | .3000 (.04) |
| 50% | .4361 (.17) | .4667 (.00) | .8016 (.01) | .7000 (.01) |
| C4.5 | .5056 | .1429 | .1984 | .0667 |

Table 12: Distribution Amount Classification Results

| Approach | Speedup |
|---|---|
| **Eureka** | 74.24 |
| Random Processor LB | 70.75 |
| Local Ordering | 68.92 |
| Transformation Ordering | 66.13 |
| Kumar and Rao | 65.89 |
| Distributed Tree | 65.82 |
| Fixed Evaluation 1 | 65.41 |
| 1 Cluster | 65.21 |
| Neighbor LB | 65.21 |
| 30% Distribution | 65.21 |
| 2 Clusters | 64.97 |
| 50% Distribution | 61.94 |
| Fixed Evaluation 2 | 49.58 |
| 4 Clusters | 49.57 |
| **Avg. of Fixed Strategies** | 63.43 |

Table 13: Combination of C4.5 Recommendations

### 5.6 Combining C4.5 Recommendations

Up to this point, all experiments have shown the results of Eureka selecting a single strategy, all other strategy results being fixed. In this experiment we allow Eureka to select all strategy choices at once for a given problem and execute the parallel search with the recommended strategies. We then compare the results to each fixed strategy (the fixed strategy choice is averaged over all problem instances and all possible choices of other strategy decisions). A random set of 50 problems from the fifteen puzzle domain is selected and run on 64 processors of the nCUBE 2. Table 13 summarizes the speedup for each approach.

These results indicate that Eureka can effectively make all strategy choices at once. The learned rules achieve better performance than that obtained by any one of these strategy choices. These rules also outperform any single fixed strategy choice averaged over all other parameter options.

| Method | Error |
|---|---|
| ID3 | 0.1067 |
| CN2 | 0.1133 |
| C4.5 | 0.1657 |
| Bayesian | 0.4471 |
| Majority | 0.4714 |
| Backprop | 0.7657 |

Table 14: Machine Learning Comparison

### 5.7 Machine Learning Comparison

In the current version of the Eureka system, we use C4.5 to induce a decision tree based on the training data. C4.5 has proven to be effective in predicting the strategy choices for these test domains. In addition, the output of the system is available as a symbolic rule base, which may allow the system developer to determine the factors that affect strategy selection for a given application domain.

Other machine learning approaches can also be used to perform strategy selection in the Eureka system. To test the results of various existing approaches, we supplied the data from all of the 15 Puzzle classification experiments described in the previous section as input to versions of C4.5, the ID3 decision tree induction algorithm (Quinlan, 1986), the CN2 sequential covering algorithm (Clark & Niblett, 1989), a backpropagation neural net (Rumelhart & McClelland, 1986), a Bayesian classifier (Cestnik, 1990), and a majority-wins classifier. As with the other experiments, results are based on ten-fold cross-validation.

Table 14 shows that the decision tree algorithms performed best on this particular data set. Ultimately, the best machine learning algorithm in this context is the algorithm that consistently yields the best speedup. If we consider normalized problem speedups, the algorithm that produces the best classification on average will also produce the greatest speedup. We will continue to explore various machine learning methods to determine the approach that will work best for this type of application.

## 6. Conclusions and Future Work

This paper reports on work performed to combine the benefits of parallel search approaches in the Eureka system. Experimentation reveals that strategies developed over the last few years offer distinct benefits to improving the performance of AI applications. However, while any particular algorithm can provide significant speedup for one type of problem, on other problems these algorithms can actually produce worse results than using a serial version of the search algorithm. As a result, these strategies need to be carefully chosen based on the characteristics of a particular problem.

In this paper we demonstrate the ability of Eureka to automatically select parallel search strategies and set appropriate parameters. Eureka employs the C4.5 machine learning system to make an intelligent choice of strategies for each new problem instance. Experiments from the domains of the fifteen puzzle problem, robot arm motion planning, an artificial search space, and planning problems indicate that Eureka yields both improved speedup results and significantly improved classification accuracies over any strategy used in isolation when high-variance training cases are used. These experiments also demonstrate Eureka's ability to select all parameters at once and to outperform any fixed strategy over a set of problem instances. In addition, we demonstrate that Eureka can benefit by utilizing training cases drawn from a variety of test domains, thus we would expect the performance of the system to improve even more as we incorporate data from new problem domains and architectures.

Two of the challenges introduced by our research are the ability to determine discriminating features and the ability to provide clear classifications for training examples. In our experiments we verified that the features we chose could be measured early during the search process and still be representative of the problem space later on during execution. As we apply our approach to a greater variety of problems we will need to develop a more formal methodology for selecting representative and discriminating features. In addition, we observed dramatic performance improvements when test cases were drawn from problem instances with clear classifications. We would like to pursue methods of learning from instances that exhibit low variation in performance of alternative strategies and high variation in problem size.

The current implementation of Eureka focuses on an IDA* approach to search. One reason for this choice of search method is the linear memory requirements of the algorithm. A second advantage of this search method is that an iterative deepening search method provides feedback in each iteration that can be used to adjust parameters for the next search iteration. As a result, Eureka can potentially adjust the strategy choices from one iteration of the search algorithm to the next as features of the space vary. However, in some problem domains non-iterative search algorithms may be preferred. A future challenge for our research is to refine the adaptive parallel search algorithm for use in a greater variety of iterative and non-iterative search algorithms.

Eureka implementations are currently available on a variety of architectural platforms including MIMD distributed memory and shared memory multiprocessors, a distributed network of machines running PVM, Posix multithreading machines, and machines using Java threads and Cilk threads. Problem domains currently under investigation include additional combinatorial optimization problems such as the n-Queens problem and integration of machine learning, theorem proving, and natural language algorithms into this search architecture. We hope to demonstrate that parallel heuristic search algorithms can yield both optimal *and* scalable approaches to planning, machine learning, natural language, theorem proving, and many other computation-intensive areas of AI.

## Acknowledgements

This work was supported by National Science Foundation grants IRI-9308308, IRI-9502260, and DMI-9413923. The authors would like to thank Dan Burns and Matthias Imhof at the MIT Earth Resources Lab for providing access to their nCUBE 2 to complete the experiments reported in this paper.